\documentclass[10pt,twocolumn,letterpaper]{article}

\usepackage{cvpr}
\usepackage{times}
\usepackage{epsfig}
\usepackage{graphicx}
\usepackage{amsmath}
\usepackage{amssymb}
\usepackage{booktabs}
\usepackage{array, multirow}
\usepackage{caption}
\graphicspath{{eps/}{jpeg/}}

\usepackage[breaklinks=true,bookmarks=false]{hyperref}

\cvprfinalcopy % *** Uncomment this line for the final submission

\def\cvprPaperID{1} % *** Enter the CVPR Paper ID here

\ifcvprfinal\fi
\begin{document}

%%%%%%%%% TITLE
\title{CASSOD-Net: Cascaded and Separable Structures of Dilated Convolution for Embedded Vision Systems and Applications}
\author{Tse-Wei Chen$^*$,
~Deyu Wang$^\dag$,
~Wei Tao$^\dag$,
~Dongchao Wen$^\dag$,
~Lingxiao Yin$^\dag$,\\
~Tadayuki Ito$^*$,
~Kinya Osa$^*$,
~and Masami Kato$^*$\\
{\tt\small twchen@ieee.org}\\
 $^*$Device Technology Development Headquarters, Canon Inc., Japan\\
 $^\dag$Canon Innovative Solution (Beijing) Co., Ltd., China\\
}

 %30-2, Shimomaruko 3-chome, Ohta-ku, Tokyo 146-8501, Japan\\
 %12A Floor, Yingu Building, No.9 Beisihuanxi Road, Haidian, Beijing, China
 
%{\tt\small twchen@ieee.org}
% For a paper whose authors are all at the same institution,
% omit the following lines up until the closing ``}''.
% Additional authors and addresses can be added with ``\and'',
% just like the second author.
% To save space, use either the email address or home page, not both
% \and
% {\tt\small secondauthor@i2.org}
%}

\maketitle
\thispagestyle{empty}

%%%%%%%%% ABSTRACT
\begin{abstract}
The field of view (FOV) of convolutional neural networks is highly related to the accuracy of inference. Dilated convolutions are known as an effective solution to the problems which require large FOVs. However, for general-purpose hardware or dedicated hardware, it usually takes extra time to handle dilated convolutions compared with standard convolutions. In this paper, we propose a network module, Cascaded and Separable Structure of Dilated (CASSOD) Convolution, and a special hardware system to handle the CASSOD networks efficiently. A CASSOD-Net includes multiple cascaded $2 \times 2$ dilated filters, which can be used to replace the traditional $3 \times 3$ dilated filters without decreasing the accuracy of inference. Two example applications, face detection and image segmentation, are tested with dilated convolutions and the proposed CASSOD modules. The new network for face detection achieves higher accuracy than the previous work with only 47\% of filter weights in the dilated convolution layers of the context module. Moreover, the proposed hardware system can accelerate the computations of dilated convolutions, and it is 2.78 times faster than traditional hardware systems when the filter size is $3 \times 3$.
\end{abstract}

%-------------------------------------------------------------------------

\begin{figure}[h]
\begin{center}
   \includegraphics[width=0.9\linewidth]{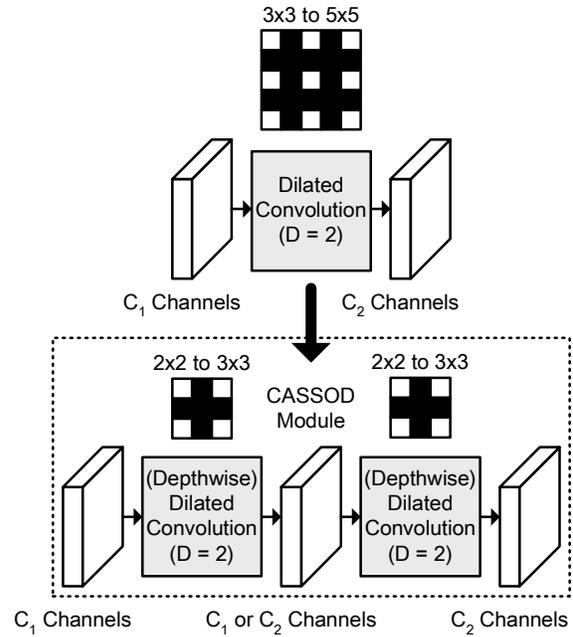}
\end{center}
   \caption{Concept of the proposed CASSOD module.}
\label{fig:concept}
\end{figure}

%%%%%%%%% BODY TEXT
\section{Introduction}
\label{sec:introduction}

Dilated convolutions in Convolutional Neural Networks (CNNs) can be applied to different kinds of applications, including audio processing~\cite{Hussain19}, crowd counting~\cite{Li18_0}, semantic image segmentation~\cite{Hamaguchi17,Mehta18,Wang18,Wu19}, image classification~\cite{Lei19}, image super-resolution~\cite{Lin18}, road extraction~\cite{Zhou18}, image denoising~\cite{Tian18}, face detection~\cite{Li18_1}, object detection~\cite{Li18_2}, and so on. For many vision applications, the receptive field, or the field of view (FOV) ~\cite{Luo17} of convolutional neural networks is highly related to the accuracy of inference. Large FOV can be obtained by increasing the filter size or by increasing the number of convolution layers. However, the computational time might also increase because the computational cost is proportional to the filter size and the number of layers. Dilated convolutions are known as an effective method to enlarge the FOV of a network without increasing the computational costs. By adjusting the dilation rate $D$, a $3 \times 3$ filter can be enlarged to a $(2D + 1) \times (2D + 1)$ filter with $3 \times 3$ Multiply-Accumulate (MAC) operations.

Deng et al. propose a context module, which is inspired by the SSH face detector~\cite{Najibi17}, to increase the FOV~\cite{Deng19}. Wu et al. propose a Joint Pyramid Up-sampling (JPU) module and formulate the task of extracting high-resolution feature maps into a joint up-sampling problem~\cite{Wu19}. In the JPU structure, the generated feature maps are up-sampled and concatenated, and 4 separable convolutions with different dilation rates ($D = 1, 2, 4, 8$) are included in the same convolution layer. Hamaguchi et al. propose a segmentation model, which includes a front-end module, a local feature extraction (LFE) module, and a head module~\cite{Hamaguchi17}. The dilation rate in the front-end module is gradually increased to extract the features from a large range, and the dilation rate in the subsequent LFE module is gradually decreased to aggregate local features generated by the front-end module. The above mentioned network models are designed for applications related to face detection and image segmentation. There are still many kinds of network architectures containing dilated convolution layers, which are used to improve the accuracy of other applications.

In order to implement dilated convolutions on mobile devices and embedded system platforms, it is necessary to reduce the memory size and the computational costs without decreasing the accuracy. For some hardware devices, including GPUs, convolution operations are only optimized for standard convolutions. Even though the $3 \times 3$ dilated filters require only $3 \times 3$ MAC operations, the overhead to skip the pixels which do not engage in the process of convolutions is not always zero. For some dedicated hardware systems based on systolic arrays, when the dilation rate is high, the memory footprint might increase since the addresses of the pixels to be processed are not consecutive.

In this paper, we propose a new network module, Cascaded and Separable Structure of Dilated (CASSOD) Convolutions, to approximate the $3 \times 3$ dilated convolutions with low memory cost for filter weights and low computational costs for convolutions. The concept of the proposed method is shown in Fig.~\ref{fig:concept}.

%The paper is organized as follows. In Sec.~\ref{sec:module}, the proposed network module, CASSOD, is introduced. The proposed hardware architecture is shown in Sec.~\ref{sec:architecture}. The experimental results are discussed in Sec.~\ref{sec:results}. The conclusions are given in Sec.~\ref{sec:conclusion}.

%-------------------------------------------------------------------------
        
\section{Proposed Network Module}
\label{sec:module}

\begin{table}
\caption{Comparison of Number of Filter Weights}
\begin{center}
\begin{tabular}{l|c}
\hline
Network Structure &  No. of Filter Weights \\
\hline
Dilated Convolution  & $3^{2} \times C_{1} \times C_{2}$\\
\hline
CASSOD-A             & \multirow{2}{*}{$2^{2} \times C_{1} \times (1 + C_{2})$}  \\
(1st layer: DW Conv.$^{*}$)\\
\hline
%CASSOD-B             & $2^{2} \times (C_{1} + 1) \times C_{2}$  \\
%(2nd layer: DW Conv.$^{*}$)\\
%\hline
\multirow{3}{*}{CASSOD-C (No DW Conv.$^{*}$)}  & $2^{2} \times (C_{1} + C_{2}) \times C_{1}$  \\
                     &  or     \\
                     & $2^{2} \times (C_{1} + C_{2}) \times C_{2}$  \\
\hline
Dilated / Depthwise Convolution  & $3^{2} \times C_{1}$\\
\hline
CASSOD-D             & \multirow{2}{*}{$2^{2} \times  (C_{1} \times 2)$}  \\
(1st and 2nd layer: DW Conv.$^{*}$)\\           
\hline
\end{tabular}
\end{center}
{\footnotesize
$^{*}$DW Conv. denotes the depthwise convolutions.\\
}
\label{tab:weights}
\end{table}

In this section, the $2 \times 2$ dilated convolutions and the proposed CASSOD module are introduced in the following subsections. The cascaded $2 \times 2$ dilated convolution, which can be a separable version of the traditional $3 \times 3$ dilated convolution, is an important component of the CASSOD module. In this paper, the $2 \times 2$ (or $3 \times 3$) dilated convolutions refer to the convolutions where the filter size is $2 \times 2$ (or $3 \times 3$) before the filters are dilated ($D = 1$).

\subsection{Dilated Convolutions with Fewer Filter Weights}
\label{subsec:dilated}

In previous works~\cite{Hamaguchi17, Hussain19, Lei19, Li18_0, Lin18, Mehta18, Tian18, Wu19, Zhou18}, the dilated filters are generated based on $3 \times 3$ filters. A total of 9 filter weights are used to compute the convolution results for one input feature map. The FOV is expanded when the dilation rate $D$ is larger than 1. The larger the dilation rate $D$, the larger the FOV.

The dilated filters which are generated based on $2 \times 2$ filters, are proposed in this work. The output result of the proposed dilated filters is calculated based on the following equation.
\begin{eqnarray}
  O_{(i,j)} = \nonumber \\
  \sum^{C-1}_{c=0} \sum^{1}_{x=0} \sum^{1}_{y=0} I_{\left(c,i - \frac{D}{2}+ x \cdot D, j- \frac{D}{2} + y \cdot D\right)}  \times W_{(c,x,y)},
\end{eqnarray}
where $O_{(i,j)}$ is the output feature map, $I_{(c,i,j)}$ is the $c$-th input feature map, and $W_{(c,x,y)}$ is the filter weight for the $c$-th input feature map. The parameters $(i,j)$ represent the index of the pixels in the feature maps, and the ranges of both $i$ and $j$ depend on the size of feature maps. The parameter $C$ denotes the number of channels. There are only 4 filter weights for 1 input feature map and the ranges of both $x$ and $y$ are $[0,1]$. 

\begin{figure}[h]
\begin{center}
   \includegraphics[width=0.7\linewidth]{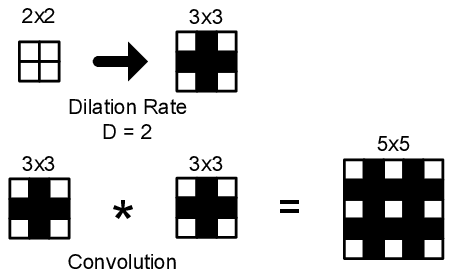}\\
   (a)\\
   \includegraphics[width=0.8\linewidth]{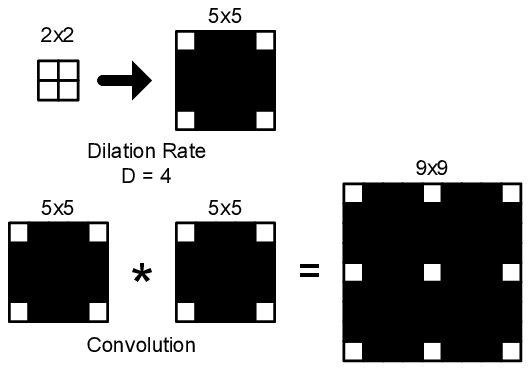}\\
   (b)\\
\end{center}
   \caption{Example of the $2 \times 2$ dilated convolutions where (a) the dilation rate $D = 2$ and (b) the dilation rate $D = 4$. }
\label{fig:dilated}
\end{figure}

Two examples of the $2 \times 2$ filters and dilated convolutions are shown in Fig.~\ref{fig:dilated}. Fig.~\ref{fig:dilated}(a) shows an example where the dilation rate $D$ is $2$. Two cascaded $2 \times 2$ dilated convolutions where the dilation rate $D$ is $2$ can be an approximation to a $3 \times 3$ dilated convolution with the same dilation rate since the positions of zero weights are exactly the same. Similarly, as shown in Fig.~\ref{fig:dilated}(b),  two cascaded $2 \times 2$ dilated convolutions with the dilation rate $D=4$ can be an approximation to a $3 \times 3$ dilated convolution with the same dilation rate. The dilation rate $D$ is always set to a multiple of 2.

\subsection{Cascaded and Separable Structure of Dilated (CASSOD) Convolution}
\label{subsec:cassod}

The concept of the proposed CASSOD module is shown in Fig.~\ref{fig:concept}. The upper part of the figure shows an example of dilated convolutions~\cite{Wei18} where the dilation rate $D$ is $2$. There are $C_{1}$ channels in the input feature maps, and $C_{2}$ channels in the output feature maps. The number of filter weights of the dilated convolutions is $3 \times 3 \times C_{1} \times C_{2}$. 

The lower part of the figure shows an example of the proposed CASSOD module where the dilation rate $D$ is $2$. There are two convolution layers in this module and 3 sets of feature maps. Either of the first convolution layer or the second convolution layer can be a depthwise convolution~\cite{Howard17} layer. The variations (CASSOD-A,C,D) are shown in Table~\ref{tab:weights}.

In the CASSOD-A module, there are $C_{1}$, $C_{1}$, and $C_{2}$ channels in the first, the second, and the third set of feature maps, respectively. The numbers of channels in the first and the second set of feature maps are the same because the first convolution layer includes depthwise separable operations. The first convolution layer includes a series of $2 \times 2$ depthwise and dilated filters, which are different from the traditional $3 \times 3$ dilated filters. The second convolution layers includes a series of $2 \times 2$ dilated filters. The number of filter weights of the dilated convolutions is $2^{2} \times C_{1}  \times (1 + C_{2})$. Table~\ref{tab:weights} also shows the comparison of the number of filter weights. It can be observed that, when $C_{2}$ is large, the number of filter weights in the CASSOD-A module is close to $44.4\%$ (4/9) of the number of filter weights in the traditional dilated convolution. 

%In the CASSOD-B module, there are $C_{1}$, $C_{2}$, and $C_{2}$ channels in the first, the second, and the third set of feature maps, respectively. The first convolution layers includes a series of $2 \times 2$ dilated filters. The second convolution layers includes a series of $2 \times 2$ depthwise and dilated filters. The number of filter weights of the dilated convolutions is $2^{2} \times (C_{1} + 1)  \times C_{2}$. Similarly, when $C_{1}$ is large, the number of filter weights in the CASSOD-B module is close to $44.4\%$ (4/9) of the number of filter weights in the traditional dilated convolution. 

In the CASSOD-C module, there are $C_{1}$, $C_{1}$ (or $C_{2}$), and $C_{2}$ channels in the first, the second, and the third set of feature maps, respectively. Both convolution layers include a series of $2 \times 2$ dilated filters. The number of filter weights of the dilated convolutions is $2^{2} \times (C_{1} + C_{2}) \times C_{1}$ or $2^{2} \times (C_{1} + C_{2}) \times C_{2}$. It can be observed that, when $C_{1}$ is equal to $C_{2}$, the number of filter weights of the CASSOD-C module is close to $88.9\%$ (8/9) of the number of filter weights in the traditional dilated convolution.

The CASSOD-A,C modules have lower computational costs than the traditional $3 \times 3$ dilated convolutions. The number of filter weights is also proportional to the computational cost, which is equal to the number of MAC operations. Compared with the traditional $3 \times 3$ dilated convolutions, 1 additional convolution layer is required to implement the CASSOD modules. The batch normalization operations and the activation functions (e.g. ReLU) can be included in the additional convolution layer if necessary. The CASSOD-A,C modules are alternatives to the dilated convolutions, and the CASSOD-D module is an alternative to the depthwise and dilated convolutions. The number of filter weights of the CASSOD-D module is close to $88.9\%$ (8/9) of the depthwise and dilated convolutions.

%-------------------------------------------------------------------------

\begin{figure}[t]
\begin{center}
   \includegraphics[width=1.0\linewidth]{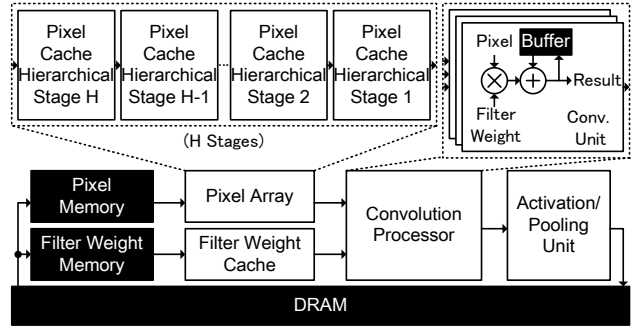}\\
\end{center}
   \caption{An example of the architecture of the proposed hardware system.}
\label{fig:system}
\end{figure}

\section{Proposed Hardware Architecture}
\label{sec:architecture}

The analysis of computational time of dilated convolutions and the proposed hardware architecture are shown in the following subsections.

\subsection{Computational Time of Dilated Convolutions}
\label{subsec:time}

Table~\ref{tab:time} shows an example of the computational time of dilated convolutions when the size of input image is $320 \times 320$ pixels. A network with 3 layers is used to measure the computational time of CPU, and a network with 10 layers is used to measure the computational time of GPU. There are 64 input channels and 64 output channels in each layer of the networks, and the dilation rates $D$ of the $3 \times 3$ convolution layers are set to the same value. 

The results show that the processing time does not change much when the dilation rate $D$ is larger than 1 for either CPU or GPU. The computational time of standard convolutions, where the dilation rate is set to 1, is the shortest, and the reason can be that the framework (e.g. CUDA library) includes some optimized operations to accelerate the computations of standard convolutions. The speed of dilated convolutions may depend on the performance of system platforms, but the dilated convolutions with a dilation rate larger than 1 cannot necessarily achieve the same level of speed as standard convolutions with a dilation rate of 1. The goal of this work is to design a hardware system which can handle both dilated convolutions and standard convolutions efficiently.

\begin{table}
\caption{Processing Time of Dilated Convolutions}
\begin{center}
\begin{tabular}{l|cc|cc}
\hline
\multirow{3}{*}{Dilation Rate} 
               & \multicolumn{4}{c}{Processing Time (ms)}\\
\cline{2-5}
               & \multicolumn{2}{c|}{Convolutions} & \multicolumn{2}{c}{Depthwise Conv.}\\ 
\cline{2-5}               
               & {CPU$^{*}$}  & GPU$^{**}$      & {CPU$^{*}$}  & GPU$^{**}$\\
              
\hline
$D = 1$ & 39.53  & 6.53  & 21.57  & 11.81\\
$D = 2$ & 862.56 & 19.12 & 861.99 & 11.83\\
$D = 3$ & 868.83 & 19.17 & 865.09 & 11.79\\
$D = 4$ & 854.23 & 19.21 & 862.95 & 11.78\\
$D = 5$ & 852.42 & 19.23 & 871.42 & 11.78\\
\hline
\end{tabular}
\end{center}
{\footnotesize
$^{*}$The CPU is Intel Xeon E5-2640 v4 (2.40 GHz) and the memory size is 256 GB.\\
$^{**}$The GPU is TITAN Xp (12 GB), and the version of cuDNN is 7.6.5.\\
}
\label{tab:time}
\end{table}

For hardware implementation, the pixel values in the feature maps are usually stored in a shift register array. One solution to handle dilated convolutions using a shift register array is to add zeros to the filter weights and compute the results of standard convolutions. To handle a dilated $3 \times 3$ filter where the dilation rate $D$ is $2$, it is necessary to compute the products of 25 filter weights and 25 input pixels. A total of 16 filter weights equal to zero can be skipped. The computational time is proportional to the size of dilated filters, and the efficiency of filter processing is relatively low compared with CPU and GPU.

To solve this problem, a new hardware architecture is proposed to speed up dilated convolutions. An example of the architecture of the proposed hardware system is shown in Fig.~\ref{fig:system}, which includes 6 modules and DRAM. The filter weights stored in the ``Filter Weight Memory" are sent to the ``Filter Weight Cache," and the feature maps of the current convolution layer stored in the ``Pixel Memory" are sent to the ``Pixel Array." The filter weights and the pixels of feature maps are sent to the ``Convolution Processor" to compute the convolution results, and the convolution results are sent to the ``Activation and Pooling Unit" to generate the feature maps for the next convolution layer. The ``Pixel Array" includes multiple hierarchical stages of pixel caches, which can generate the input pixels for dilated convolutions with a different dilation rate $D$.

\subsection{Pixel Array for Dilated Convolutions}
\label{subsec:shift}

\begin{figure*}[t]
\begin{center}
   \includegraphics[width=0.75\linewidth]{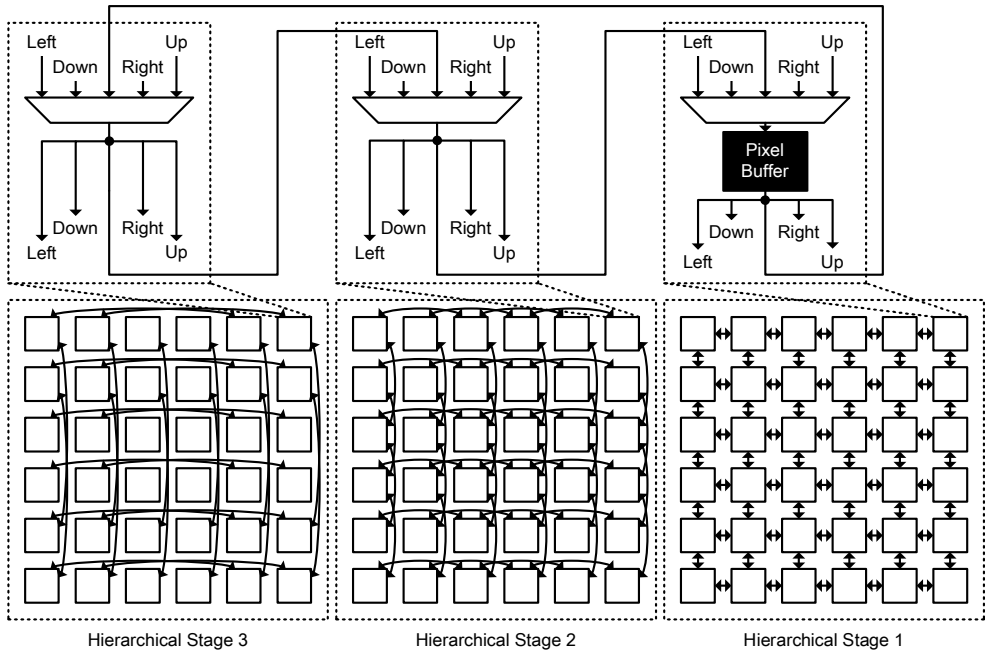}\\
\end{center}
   \caption{An example of the hardware architecture and the interconnections of the ``Pixel Array" with $6 \times 6$ pixels.}
\label{fig:shift}
\end{figure*}

The ``Pixel Array," including multiple hierarchical stages, can handle dilated convolutions with different dilation rates efficiently. The proposed hardware architecture can be implemented in a Field-Programmable Gate Array (FPGA) or an Application-Specific Integrated Circuit (ASIC). Fig.~\ref{fig:shift} shows an example of the architecture and the interconnections of the proposed shift register array, where there are 3 hierarchical stages. In this example, there are $6 \times 6$ pixel buffers (or selectors) in each stage, and each pixel buffer (or selector) is connected to 4 other pixel buffers (or selectors) in the up, right, down, and left directions. It means that the pixel stored in the pixel buffer can be transferred to one of the 4 connected pixel buffers (or selectors). In the first hierarchical stage, the input of a pixel buffer is connected to the input of its neighboring pixel buffer and the pixel selector with the same position in the second stage. The difference of the shifted index and the original index of the pixel, $X_{0}$, is 0 or 1 ($2^0$). Similarly, in the second hierarchical stage, the difference of the shifted index and the original index of the pixel, $X_{1}$, is 0 or 2 ($2^1$), and in the third hierarchical stage, the difference of the shifted index and the original index of the pixel, $X_{2}$, is 0 or 4 ($2^2$).

The supported dilation rate, which is equal to the total difference of the shifted index and the original index of the pixel in all hierarchical stages ($X_{0}$, $X_{1}$, $X_{2}$, ... , $X_{H-1}$) is shown in the following equation.
\begin{eqnarray}
  D=\sum^{H-1}_{h=0} X_{h} \cdot 2^{h},
\label{eq:supported}
\end{eqnarray}
where $H$ is the total number of the hierarchical stages. In this example, $H$ is equal to 3, but $H$ can be set to any number to support $2^H - 1$ dilation rates ($D=[1, 2^{H}-1]$) according to the network architectures. When the gate count of each stage is the same, the total hardware cost is proportional to the number of hierarchical stage, $H$.

%-------------------------------------------------------------------------
\section{Experimental Results}
\label{sec:results}

The experimental results contain 3 parts. The first part is the comparison of accuracy of face detection. The second part is the comparison of accuracy of image segmentation. The third part is the analysis of the performance of the proposed hardware architecture.

%-------------------------------------------------------------------------

\subsection{Accuracy of Face Detection}
\label{subsec:face}

\begin{figure*}[t]
\begin{center}
   \includegraphics[width=0.7\linewidth]{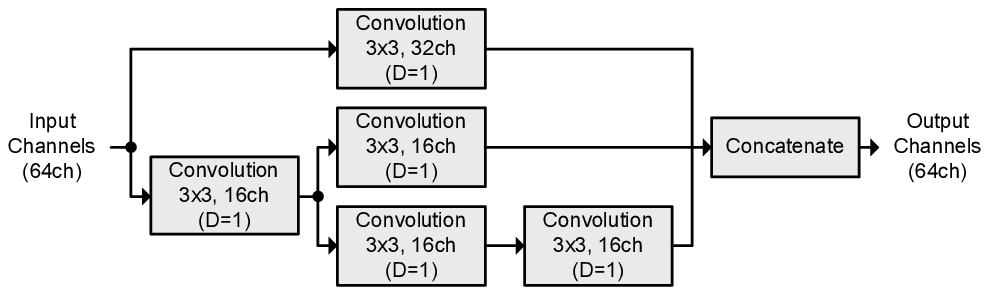} (a)\\
   \includegraphics[width=0.7\linewidth]{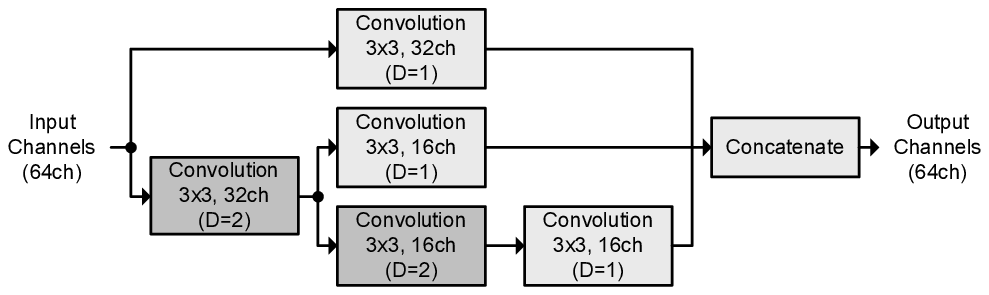} (b)\\
\end{center}
   \caption{(a) The context module of RetinaFace~\cite{Deng19} with the SSH structure~\cite{Najibi17} and (b) the modified context module with the FEM structure~\cite{Li18_1}, in which the dilated convolutions can be replaced by the proposed CASSOD modules.}
\label{fig:dsfd}
\end{figure*}

The proposed network architecture, CASSOD-Net, is evaluated based on the RetinaFace~\cite{Deng19}, which is designed for face detection tasks. The context module, inspired by the SSH face detector~\cite{Najibi17}, is used to increase the FOV and enhance the rigid context modeling power. To compare the accuracy, the convolution layers in the context modules is replaced by the Feature Enhance Module (FEM), which  includes dilated convolution structures, in the Dual Shot Face Detector (DSFD)~\cite{Li18_1}. Then, the dilated convolutions in the DSFD~\cite{Wu19} are replaced by the proposed CASSOD modules. The network architecture of the modified context module is shown in Fig.~\ref{fig:dsfd}. To keep the computational costs at the same level, the number of input channels of the FEM is changed from 256 to 64, and the number of input channels in the remaining layers is reduced by the same ratio. Fig.~\ref{fig:dsfd}(a) shows the original context module in the RetinaFace, and Fig.~\ref{fig:dsfd}(b) shows the modified context module with FEM, which includes dilated convolution layers with a dilation rate ($D$) of 2.

MobileNet-0.25~\cite{Howard17} is used as the backbone network. The accuracy of face detection is shown in Table~\ref{tab:dsfd}. After replacing the dilated convolutions with the CASSOD-C and CASSOD-A modules, the accuracy does not decrease. Besides, the proposed CASSOD module can achieve higher accuracy than the previous architecture with batch normalization and activation. The CASSOD-C module has better performance than the CASSOD-A module on the 3 categories (easy, medium, and hard), and the reason can be that the CASSOD-C module has more parameters than the CASSOD-A module. It is shown that the CASSOD module is a good alternative to the original dilated convolutions. The accuracy of the CASSOD-A module shows that the new network achieves higher accuracy than the FEM-based network with only 47\% of filter weights in the dilated convolution layers of the context module.

\begin{table*}
\caption{Comparison of Accuracy of Face Detection~\cite{Deng19} between Dilated Convolutions and CASSOD Modules}
\begin{center}
\begin{tabular}{l|ccc|c}
\hline
  \multirow{2}{*}{Context Module of RetinaFace~\cite{Deng19}}    & \multirow{2}{*}{Easy (\%)}   & \multirow{2}{*}{Medium (\%)}    & \multirow{2}{*}{Hard (\%)}          & Parameter Size of     \\
                                                                 &                              &                                 &                                     & Dilated Conv. Layers  \\
\hline
  SSH$^{\dag}$                                                   & 88.72                        & 86.97                           & 79.19                                & (11,520)          \\
\hline
  FEM$^{\dag \dag}$                                              & 88.87                        & 86.74                           & 80.26                                &  23,040               \\
  FEM$^{\dag \dag}$-CASSOD-C                                     & 89.05                        & 87.46                           & 81.09                                &  15,360               \\
  FEM$^{\dag \dag}$-CASSOD-C with BN$^{*}$                       & \textbf{89.21}               & 87.55                           & \textbf{81.28}                       &  15,552               \\
  FEM$^{\dag \dag}$-CASSOD-C with BN$^{*}$ and ReLU$^{**}$       & 89.12                        & \textbf{87.62}                  & 81.16                                &  15,552               \\                
  FEM$^{\dag \dag}$-CASSOD-A with BN$^{*}$ and ReLU$^{**}$       & 88.88                        & 87.40                           & 80.74                                &  \textbf{10,912}      \\                
\hline
\end{tabular}
\end{center}
{\footnotesize
$^{\dag}$SSH represents single stage headless face detector~\cite{Najibi17}. The parameter size of convolution layers to be replaced by dilated convolution layers and CASSOD modules is 11,520.\\
$^{\dag \dag}$FEM represents feature enhance module~\cite{Li18_1}.\\
$^{*}$BN represents batch normalization.\\
$^{**}$ReLU represents rectified linear unit.\\
}
\label{tab:dsfd}
\end{table*}

%-------------------------------------------------------------------------

\subsection{Accuracy of Image Segmentation}
\label{subsec:segmentation}

The proposed network architecture, CASSOD-Net, is evaluated based on the FastFCN~\cite{Wu19}, which is designed for semantic image segmentation tasks. The JPU module, which is included in the FastFCN, combines the up-sampled feature maps by using 4 groups of dilated convolutions with different dilation rates. The JPU modules also contain depthwise convolution layers. It can extract multiple-scale context information and increase the accuracy of image segmentation.

\begin{figure*}[t]
\begin{center}
   \includegraphics[width=0.8\linewidth]{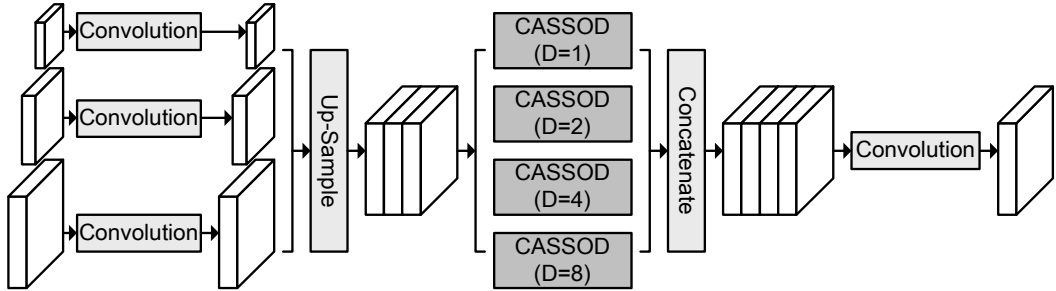}
\end{center}
   \caption{The modified JPU~\cite{Wu19} in which the dilated convolutions are replaced by the proposed CASSOD modules.}
\label{fig:jpu}
\end{figure*}

To compare the accuracy, the dilated convolutions in the JPU~\cite{Wu19} are replaced by the proposed CASSOD-D modules. The network architecture of the modified JPU is shown in Fig.~\ref{fig:jpu}. ResNet-50~\cite{He15} is used as a backbone network and trained with Pascal Context~\cite{Mottaghi14} and ADE20K datasets~\cite{Zhou17}. The accuracy of image segmentation is shown in Table~\ref{tab:jpu}. The results~\cite{Wu19} and our re-implementation results are very close. After replacing the dilated convolutions with the CASSOD modules, the accuracy does not decrease. Besides, the proposed CASSOD module can achieve higher accuracy than the previous architecture with batch normalization and activation. It is shown that the CASSOD module is a good alternative to the original dilated convolutions for image segmentation. Also, the proposed CASSOD modules can be applied to networks with depthwise and dilated convolutions. The computational time of GPU increases after applying the CASSOD modules because an extra layer is added. This problem can be solved by using the proposed hardware architecture.

\begin{table*}
\caption{Comparison of Accuracy of Image Segmentation \cite{Wu19} between Dilated Convolutions and CASSOD Modules}
\begin{center}
\begin{tabular}{l|c|ccc}
\hline
Datasets                                          &  Networks                                                & pixAcc (\%)                     & mIoU (\%)                           & Speed (fps$^{\dag}$)   \\
\hline                                                                                                                                                                                
\multirow{7}{*}{Pascal Context~\cite{Mottaghi14}} &  EncNet + JPU (Table 2, 3 in~\cite{Wu19})                &                         --      &                         51.20               & 37.56 \\
\cline{2-5}
                                                  &  EncNet + JPU (Re-implementation)                        &                         77.88   &                         49.44               & \textbf{35.30} \\
                                                  &  EncNet + JPU-CASSOD-D (Proposed Work)                   &                         79.52   &                         52.51               & 34.73 \\
                                                  &  EncNet + JPU-CASSOD-D with BN and ReLU                  &         \multirow{2}{*}{79.67}  &         \multirow{2}{*}{52.72}              & \multirow{2}{*}{34.57}\\
                                                  & (Proposed Work)                                          &                                 &                                             &  \\
                                                  &  EncNet + JPU-CASSOD-D with BN                           & \multirow{2}{*}{\textbf{79.75}} & \multirow{2}{*}{\textbf{52.76}}             & \multirow{2}{*}{34.12} \\
                                                  & (Proposed Work)                                          &                                 &                                             &\\                                                                       
\hline                                                                                                                                                                               
\multirow{7}{*}{ADE20K~\cite{Zhou17}}             &  EncNet + JPU (Table 2, 4 in~\cite{Wu19})                &                         80.39   &                         42.75               & 37.56 \\
\cline{2-5}
                                                  &  EncNet + JPU (Re-implementation)                        &                         80.04   &                         42.09               & \textbf{35.30} \\
                                                  &  EncNet + JPU-CASSOD-D (Proposed Work)                   &                         80.35   &                         42.72               & 34.73 \\
                                                  &  EncNet + JPU-CASSOD-D with BN and ReLU                  & \multirow{2}{*}{\textbf{80.48}} & \multirow{2}{*}{\textbf{42.86}}             & \multirow{2}{*}{34.57} \\
                                                  & (Proposed Work)                                          &                                 &                                             & \\                
                                                  &  EncNet + JPU-CASSOD-D with BN                           &         \multirow{2}{*}{80.42}  &         \multirow{2}{*}{42.78}              & \multirow{2}{*}{34.12} \\
                                                  & (Proposed Work)                                          &                                 &                                             &\\
\hline
\end{tabular}
\end{center}
{\footnotesize
$^{\dag}$Fps represents frame per second, which is the unit of processing speed of the GPU, TITAN Xp (12 GB).\\
}
\label{tab:jpu}
\end{table*}

%-------------------------------------------------------------------------
\subsection{Analysis of Hardware Systems}
\label{subsec:hardware}

\begin{figure*}[t]
\begin{center}
\begin{tabular}{ccc}
\includegraphics[height=0.22\linewidth]{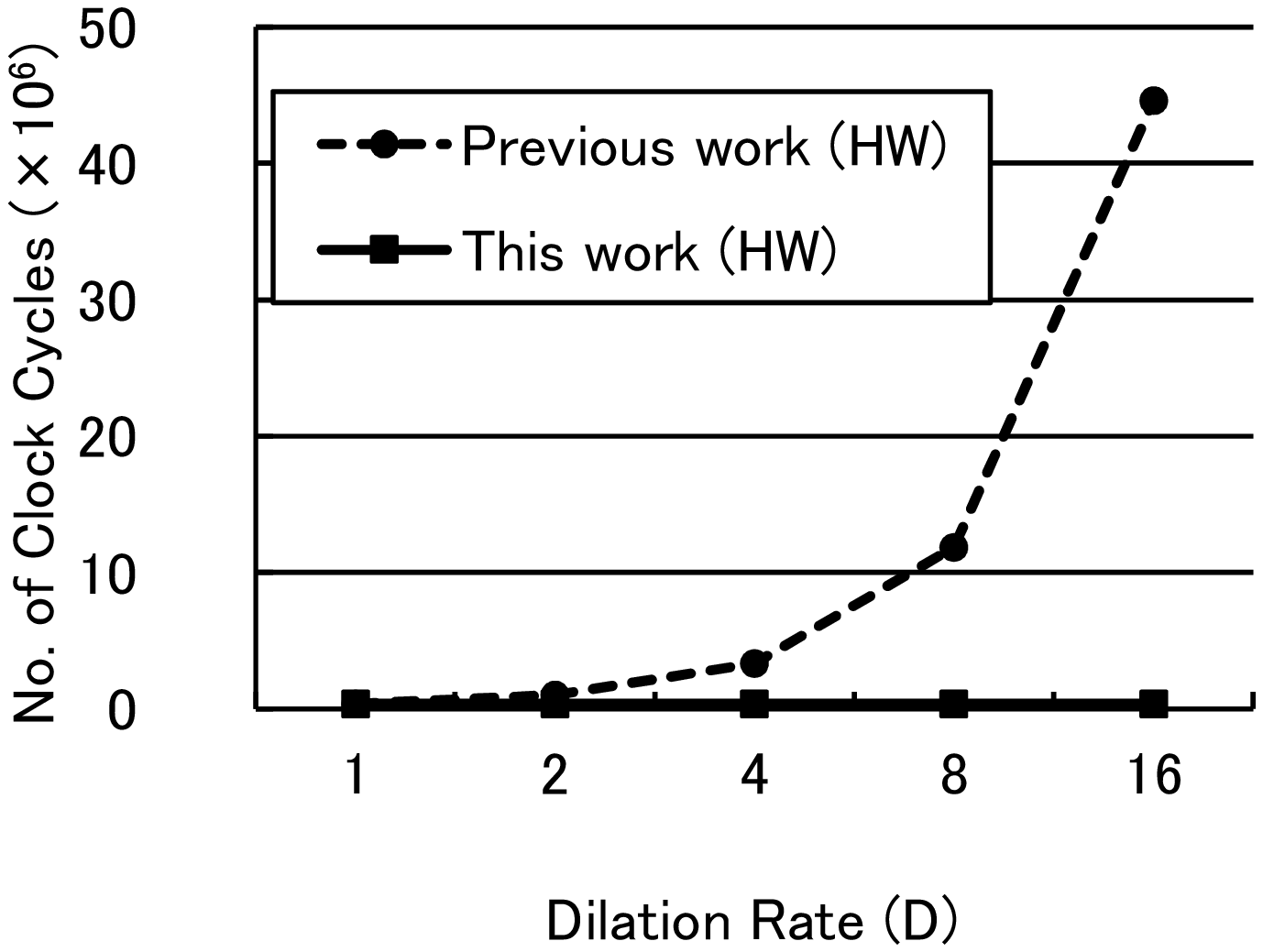} & \includegraphics[height=0.22\linewidth]{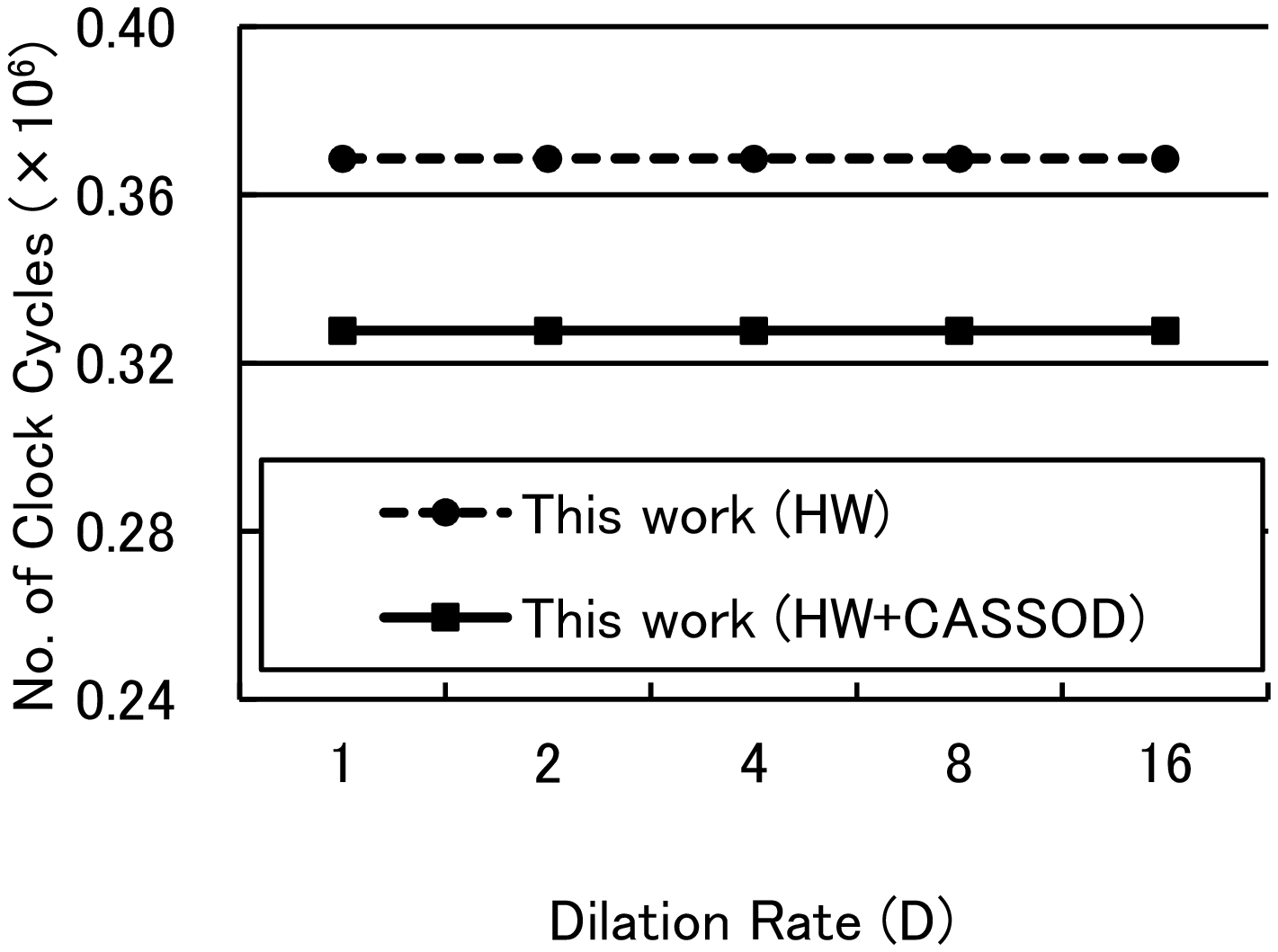} & \includegraphics[height=0.22\linewidth]{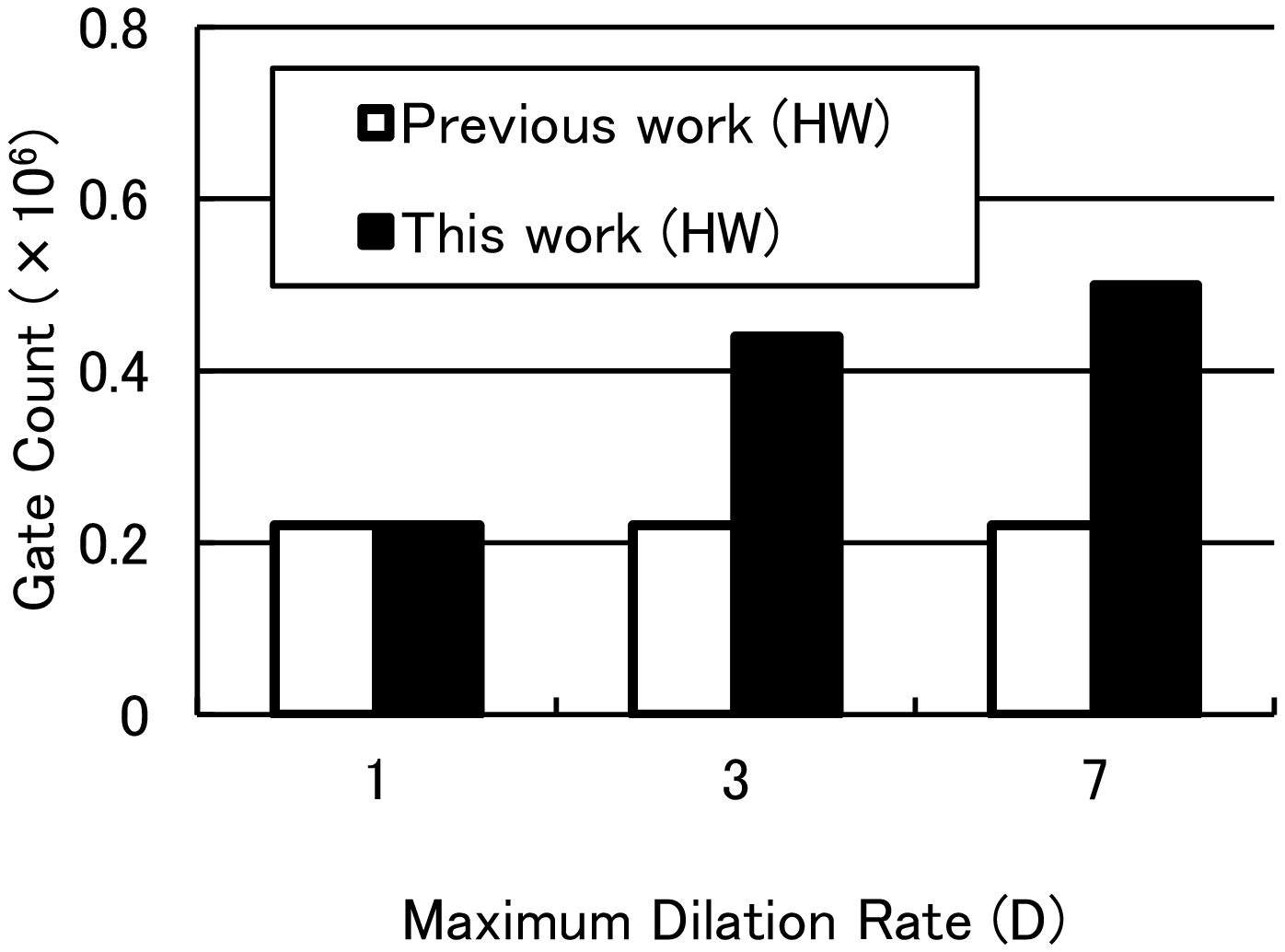} \\
(a) & (b) & (c)\\
\end{tabular}   
\end{center}
   \caption{Analysis of (a) the computational time of $3 \times 3$ filters and the dilation rate ($D$), (b) the computational time of the proposed CASSOD module and the dilation rate ($D$), and (c) the gate count of the ``Pixel Array" and the maximum dilation rate ($D$). The computational time of 1 convolution layer is measured. The size of input feature maps is $128 \times 128$ pixels, and the numbers of input channels and output channels are both set to 64. ``HW" in the figure caption represents hardware.}
\label{fig:hardware}
\end{figure*}

To show the advantages of the proposed hardware system, the hardware architecture and the CASSOD modules are compared with previous works. The previous works refer to the hardware systems~\cite{Chen16, Ma18} which do not accelerate the algorithms with the $3 \times 3$ dilated convolutions shown in Sec.~\ref{sec:module}. Since the hardware systems of the related works are not available, we re-implement a hardware system on our platforms without using the ``Pixel Array" shown in Sec.~\ref{subsec:shift} for comparison. The results are shown in Fig.~\ref{fig:hardware}.

Fig.~\ref{fig:hardware}(a) shows the relation between the computational time and the dilation rate ($D$) with $3 \times 3$ filters. The number of cycles is equivalent to the computational time. In previous works, since there are no special hardware architectures to handle dilated convolutions, it is necessary to add zero-values to the filter weights after a filter is dilated. The computational time increases as the dilation rate $D$ increases and is roughly proportional to a square of the dilation rate $(D)$. In the proposed hardware system, the input pixels of dilated convolutions can be adjusted according to the dilation rate $(D)$  and dumped consecutively, and the computational time does not vary with the dilation rate $D$. It can be observed that the proposed hardware system can handle both dilated convolutions and standard convolutions ($D = 1$) efficiently. When the dilation rate ($D$) of a $3 \times 3$ filter is 2, the proposed hardware system is 2.78 times faster than the previous work.

Fig.~\ref{fig:hardware}(b) shows the relation between the computational time and the dilation rate ($D$). By replacing the traditional dilated convolutions with the CASSOD module, the computational cost and the parameter size can be reduced even though 1 additional convolution layer is required. In the proposed hardware system, the time to set the parameters for 1 additional layer is relatively small. The results show that the computational time of the proposed hardware system can be further reduced by using the CASSOD module, which achieves similar accuracy as the traditional dilated convolutions.

Fig.~\ref{fig:hardware}(c) shows the relation between the gate count of the ``Pixel Array" and the maximum dilation rate ($D$). It can be observed that the overhead of hardware cost increases as the dilation rate $D$ increases, but the difference of hardware cost between the previous work and the proposed work is not proportional to the maximum dilation rate, $D$. The proposed hardware system scales well because it can support $2^3 - 1$ dilation rates ($D=[1, 7]$) with less than $3$ times of hardware costs.

The supported dilation rates also depend on the filter size and the interface between modules. Table~\ref{tab:spec} shows the specifications of the proposed hardware architecture. Different from previous works, the proposed hardware system can handle $2 \times 2$ and $3 \times 3$ dilated convolutions efficiently. The maximum supported dilation rate for $2 \times 2$ filters is 6, and the maximum supported dilation rate for $3 \times 3$ filters is 3. By replacing the traditional dilated convolutions with the CASSOD module, an approximated $3 \times 3$ dilated filters with a dilation rate of 6 can be implemented with cascaded $2 \times 2$ filters with a dilation rate of 6. The computational costs and memory footprints can also be reduced. Since the ''Pixel Array" with 3 hierarchical stages only occupies only 21\% of the total area, the overhead to support the dilated convolutions is relatively small. A comparison of the processing speed of the proposed system is shown in Table~\ref{tab:speed}. The resolution of the input image is $512 \times 512$ pixels, and the network is RetinaFace~\cite{Deng19} with FEM~\cite{Li18_1}, which is shown in Table~\ref{tab:dsfd}. By using the proposed hardware architecture with the shift register array, 23\% of processing time can be reduced. By replacing the dilated $3 \times 3$ filters with the proposed CASSOD modules, 9\% of processing time can be further reduced. The result clearly shows the advantages in terms of computational speed.

\begin{table}
\caption{Specifications of Proposed Hardware Architecture}
\begin{center}
\begin{tabular}{l|cc}
\hline
Gate Count                 & Total: 2.4 M \\
(NAND-Gates)               & (Pixel Array: 0.5 M)\\
\hline                     
Process                    & 28-nm CMOS technology\\
\hline                     
Clock Frequency            & 400 MHz\\
\hline                     
Supported Filter Size      & Maximum $7 \times 7$\\
Supported Dilation Rate    & $2 \times 2$ filter: $D=2, 4, 6$\\
                           & $3 \times 3$ filter: $D=1, 2, 3$\\
No. of Hierarchical Stages & $H = 3$\\
\hline
Memory Size                & 128 KB\\
\hline                     
Performance                & 409.6 GOPS*\\
\hline
\end{tabular}
\end{center}
{\footnotesize
$^{*}$GOPS represents giga operations per second. There are 2 operations in 1 MAC operation.
}
\label{tab:spec}
\end{table}
%calculations of GOPS: 512 x 0.4 x 2 = 409.6

\begin{table}
\caption{Processing Speed of Proposed Hardware}
\begin{center}
\begin{tabular}{l|cc}
\hline
                    & RetinaFace~\cite{Deng19} & RetinaFace~\cite{Deng19} \\
                    & +FEM~\cite{Li18_1}       & +FEM~\cite{Li18_1} \\
                    &                          & +CASSOD-A \\
\hline                     
without Pixel Array & 171 fps & 224 fps\\
\hline
with Pixel Array    & 222 fps & \textbf{244} fps\\
\hline
\end{tabular}
\end{center}
\label{tab:speed}
\end{table}

%-------------------------------------------------------------------------

\section{Conclusions and Future Work}
\label{sec:conclusion}

In this paper, we propose an efficient module, which is called Cascaded and Separable Structure of Dilated (CASSOD) Convolutions, and a special hardware system to handle the CASSOD networks efficiently. 

To analyze the accuracy of algorithms, two example applications, face detection and image segmentation, are tested with dilated convolutions and the proposed alternatives. For face detection, the RetinaFace~\cite{Deng19} network architecture can achieve the same level of accuracy after replacing the dilated convolutions in the context module with the proposed CASSOD modules. For image segmentation, the FastFCN~\cite{Wu19} can achieve the same level of accuracy after replacing the dilated convolutions in the JPU with the proposed CASSOD modules, which also contain depthwise convolutions. It is shown that the CASSOD module is a good alternative to the traditional dilated convolutions for both applications.

The performance of hardware is analyzed in terms of computational time and hardware costs. The input pixels of dilated convolutions can be adjusted according to the dilation rate $(D)$  and dumped consecutively, the computational time does not vary with the dilation rate $D$. By using the proposed hardware architecture with the shift register array, 23\% of processing time can be reduced for face detection applications.

The experiments clearly show that both the proposed hardware system and the proposed CASSOD modules have advantages over previous works. For future works, we plan to test the proposed hardware system and the CASSOD modules with other applications.

{\small
\bibliographystyle{ieee_fullname}
\bibliography{IEEEfull,cvprws2021_paper}

\begin{thebibliography}{10}\itemsep=-1pt

\bibitem{Chen16}
Yu-Hsin Chen, Joel Emer, and Vivienne Sze.
\newblock Eyeriss: A spatial architecture for energy-efficient dataflow for
  convolutional neural networks.
\newblock In {\em Proceedings of ACM/IEEE International Symposium on Computer
  Architecture (ISCA)}, June 2016.

\bibitem{Deng19}
Jiankang Deng, Jia Guo, Yuxiang Zhou, Jinke Yu, Irene Kotsia, and Stefanos
  Zafeiriou.
\newblock {RetinaFace}: Single-stage dense face localisation in the wild, 2019.
\newblock CoRR, abs/1905.00641.

\bibitem{Hamaguchi17}
Ryuhei Hamaguchi, Aito Fujita, Keisuke Nemoto, Tomoyuki Imaizumi, and Shuhei
  Hikosaka.
\newblock Effective use of dilated convolutions for segmenting small object
  instances in remote sensing imagery, 2017.
\newblock CoRR, abs/1709.00179.

\bibitem{He15}
Kaiming He, Xiangyu Zhang, Shaoqing Ren, and Jian Sun.
\newblock Deep residual learning for image recognition, 2015.
\newblock CoRR, abs/1512.03385.

\bibitem{Howard17}
Andrew~G. Howard, Menglong Zhu, Bo Chen, Dmitry Kalenichenko, Weijun Wang,
  Tobias Weyand, Marco Andreetto, and Hartwig Adam.
\newblock {MobileNets}: Efficient convolutional neural networks for mobile
  vision applications, 2017.
\newblock CoRR, abs/1704.04861.

\bibitem{Hussain19}
Shehzeen Hussain, Mojan Javaheripi, Paarth Neekhara, Ryan Kastner, and Farinaz
  Koushanfar.
\newblock {FastWave}: Accelerating autoregressive convolutional neural networks
  on {FPGA}.
\newblock In {\em Proceedings of IEEE/ACM International Conference on
  Computer-Aided Design (ICCAD)}, 2019.

\bibitem{Lei19}
Xinyu Lei, Hongguang Pan, and Xiangdong Huang.
\newblock A dilated {CNN} model for image classification.
\newblock {\em IEEE Access}, 7:124087--124095, July 2019.

\bibitem{Li18_1}
Jian Li, Yabiao Wang, Changan Wang, Ying Tai, Jianjun Qian, Jian Yang, Chengjie
  Wang, Jilin Li, and Feiyue Huang.
\newblock {DSFD}: Dual shot face detector, 2018.
\newblock CoRR, abs/1810.10220.

\bibitem{Li18_0}
Yuhong Li, Xiaofan Zhang, and Deming Chen.
\newblock {CSRNet}: Dilated convolutional neural networks for understanding the
  highly congested scenes, 2018.
\newblock CoRR, abs/1802.10062.

\bibitem{Li18_2}
Zeming Li, Chao Peng, Gang Yu, Xiangyu Zhang, Yangdong Deng, and Jian Sun.
\newblock {DetNet}: A backbone network for object detection, 2018.
\newblock CoRR, abs/1804.06215.

\bibitem{Lin18}
Guimin Lin, Qingxiang Wu, Lida Qiu, and Xixian Huang.
\newblock Image super-resolution using a dilated convolutional neural network.
\newblock {\em Neurocomputing}, 275:1219--1230, 2018.

\bibitem{Luo17}
Wenjie Luo, Yujia Li, Raquel Urtasun, and Richard Zemel.
\newblock Understanding the effective receptive field in deep convolutional
  neural networks, 2017.
\newblock CoRR, abs/1701.04128.

\bibitem{Ma18}
Yufei Ma, Naveen Suda, Yu Cao, Sarma Vrudhula, and Jae sun Seo.
\newblock {ALAMO}: {FPGA} acceleration of deep learning algorithms with a
  modularized {RTL} compiler.
\newblock {\em Integration}, 62:14--23, 2018.

\bibitem{Mehta18}
Sachin Mehta, Mohammad Rastegari, Anat Caspi, Linda Shapiro, and Hannaneh
  Hajishirzi.
\newblock {ESPNet}: Efficient spatial pyramid of dilated convolutions for
  semantic segmentation.
\newblock In {\em Proceedings of European Conference on Computer Vision
  (ECCV)}, 2018.

\bibitem{Mottaghi14}
Roozbeh Mottaghi, Xianjie Chen, Xiaobai Liu, Nam-Gyu Cho, Seong-Whan Lee, Sanja
  Fidler, Raquel Urtasun, and Alan Yuille.
\newblock The role of context for object detection and semantic segmentation in
  the wild.
\newblock In {\em Proceedings of IEEE/CVF Conference on Computer Vision and
  Pattern Recognition (CVPR)}, 2014.

\bibitem{Najibi17}
Mahyar Najibi, Pouya Samangouei, Rama Chellappa, and Larry Davis.
\newblock {SSH}: Single stage headless face detector, 2017.
\newblock CoRR, abs/1708.03979.

\bibitem{Tian18}
Chunwei Tian, Yong Xu, Lunke Fei, Junqian Wang, Jie Wen, and Nan Luo.
\newblock Enhanced {CNN} for image denoising, 2018.
\newblock CoRR, abs/1810.11834.

\bibitem{Wang18}
Zhengyang Wang and Shuiwang Ji.
\newblock Smoothed dilated convolutions for improved dense prediction, 2018.
\newblock CoRR, abs/1808.08931.

\bibitem{Wei18}
Yunchao Wei, Huaxin Xiao, Honghui Shi, Zequn Jie, Jiashi Feng, and Thomas~S.
  Huang.
\newblock Revisiting dilated convolution: A simple approach for weakly- and
  semi- supervised semantic segmentation, 2018.
\newblock CoRR, abs/1805.04574.

\bibitem{Wu19}
Huikai Wu, Junge Zhang, Kaiqi Huang, Kongming Liang, and Yizhou Yu.
\newblock {FastFCN}: Rethinking dilated convolution in the backbone for
  semantic segmentation, 2019.
\newblock CoRR, abs/1903.11816.

\bibitem{Zhou17}
Bolei Zhou, Hang Zhao, Xavier Puig, Sanja Fidler, Adela Barriuso, and Antonio
  Torralba.
\newblock Scene parsing through {ADE20K} dataset.
\newblock In {\em Proceedings of IEEE/CVF Conference on Computer Vision and
  Pattern Recognition (CVPR)}, 2017.

\bibitem{Zhou18}
Lichen Zhou, Chuang Zhang, and Ming Wu.
\newblock {D-LinkNet}: {LinkNet} with pretrained encoder and dilated
  convolution for high resolution satellite imagery road extraction.
\newblock In {\em Proceedings of IEEE/CVF Conference on Computer Vision and
  Pattern Recognition Workshops (CVPRW)}, 2018.

\end{thebibliography}
}

\end{document}